%% file: main.tex
\documentclass[conference]{IEEEtran}
\usepackage[latin9]{inputenc}
\usepackage{amsmath}
\usepackage{graphicx}
\usepackage{multirow}
\usepackage{multicol}
\usepackage{booktabs}
\usepackage{adjustbox}
\usepackage{xcolor}
\usepackage{xspace}
\usepackage{float}
\usepackage{subcaption}
\usepackage[hidelinks]{hyperref}

\title{Differential Anomaly Detection for Facial Images}
\author{\IEEEauthorblockN{M. Ibsen$^1$, L. J. Gonzalez-Soler$^1$, C. Rathgeb$^1$, P. Drozdowski$^1$, M. Gomez-Barrero$^2$, C. Busch$^1$}
\IEEEauthorblockA{1 - Biometrics and Internet Security Research Group \\ Hochschule Darmstadt, Germany \\
\{mathias.ibsen,lazaro-janier.gonzalez-soler,christian.rathgeb,pawel.drozdowski,christoph.busch\}@h-da.de}
\IEEEauthorblockA{2 - Hochschule Ansbach, Germany \\
marta.gomez-barrero@hs-ansbach.de}
}

\include{commands}

\begin{document}
\maketitle

\begin{abstract}
Due to their convenience and high accuracy, face recognition systems are widely employed in governmental and personal security applications to automatically recognise individuals. Despite recent advances, face recognition systems have shown to be particularly vulnerable to identity attacks (\ie, digital manipulations and attack presentations). Identity attacks pose a big security threat as they can be used to gain unauthorised access and spread misinformation. In this context, most algorithms for detecting identity attacks generalise poorly to attack types that are unknown at training time. To tackle this problem, we introduce a differential anomaly detection framework in which deep face embeddings are first extracted from pairs of images (\ie, reference and probe) and then combined for identity attack detection. The experimental evaluation conducted over several databases shows a high generalisation capability of the proposed method for detecting unknown attacks in both the digital and physical domains. 
\end{abstract}


\section{Introduction}
Face recognition systems have evolved considerably in recent years, and state-of-the-art approaches have shown impressive recognition capabilities. However, it has been shown that these systems are vulnerable to \emph{attack presentations} (APs) \cite{ISO-IEC-30107-3-PAD-metrics-170227, Raghavendra-FacePAD-Survey-2017} and \emph{digital manipulations} that induce alterations to the appearance of a face \cite{TolosanaDeepFakesBeyondSurveyOfFaceManipulationFaceDetection2020,Rathgeb-BeautificationOverview-IEEE-2019, Scherhag-MorphingSurvey-IEEE-2019}. Herein, we refer to APs and digital face manipulations jointly as \emph{identity attacks}. In those identity attacks, the facial appearance of the capture subject is altered, for instance, with the aim of concealment or impersonation. Common identity attacks include silicone masks~\cite{Kotwal-MultispectralDeepEmbeddingsCountermeasureToSiliconeMaskPAs-TBIOM-2019}, applying makeup over the face with the aim of concealment~\cite{Drozdowski-MakeupPADataset-IWBF-2021}, or manipulating facial attributes through morphing tools~\cite{Scherhag-FaceMorphingAttacks-TIFS-2020}. Identity attacks can be used to gain access to a secure application such as unlocking a smartphone  or to circumvent border controls \cite{Raghavendra-FacePAD-Survey-2017, Scherhag-MorphingSurvey-IEEE-2019}, in which face recognition systems are commonly deployed. Identity attacks are not only a concern from a security point-of-view and when used in biometric systems, but also in general media forensics, since digital identity attacks can be used to spread misinformation (\eg, the so-called DeepFakes) \cite{Verdoliva-DeepFakeOverview-ISP-2020, TolosanaDeepFakesBeyondSurveyOfFaceManipulationFaceDetection2020}. To address the aforementioned concerns, several identity attack detection techniques have been proposed in the literature~\cite{Arashloo-AnomalyDetection-IEEE-2017,Deb-UniFAD-Arxiv-2021,GonzalezSoler-PAD-FvEnc-BIOSIG-2020,GonzalezSoler-UnkownAttacksFace-IET-Biometrics-2021,Wang-PAD-AdvDomainAdaptation-TIFS-2020}. Generally, most of those approaches consider attack detection as a bi-class classification problem where classifiers are trained on both \emph{bona fide presentations} (BPs) and attack presentations. Such detection algorithms struggle to generalise well beyond the attack types that they were trained on~\cite{Khodabakhsh-FakeFaceDetection-BIOSIG-2018, Cozzolino-IDReveal-arxiv-2021}. In order to improve the generalisation capability to unknown attacks, recent studies have explored different novel approaches such as domain adaptation~\cite{Wang-PAD-AdvDomainAdaptation-TIFS-2020}, semantic hidden information through generative models~\cite{GonzalezSoler-PAD-FvEnc-BIOSIG-2020,GonzalezSoler-UnkownAttacksFace-IET-Biometrics-2021}, and anomaly detection~\cite{Arashloo-AnomalyDetection-IEEE-2017, Nikisins-OnEffectivenessOfAnamolyDetectionApproachesAgainstUnseenPresentationAttacksInFaceAntiSpoofing-ICB-2018,Fatemifar-SpoofingAttackDetectionByAnomalyDetection-2019}. Additionally, identity-aware detection methods have been proposed. In~\cite{Chingovska-OnTheUseOfClientIndentityInformationForFaceAntispoofing-IEEE-2015} the authors showed that using identity information during detection of APs could improve detection accuracy compared to approaches which did not use identity information. In \cite{Cozzolino-IDReveal-arxiv-2021} the authors created an identity-aware DeepFake video detection algorithm which detects digital manipulations in videos by learning information specific to the identity of a subject. In this context, differential detection algorithms are a type of identity-aware detection technique where both a trusted and suspected image are used during detection. Differential detection is possible since at the time of authentication pairs of images (reference and probe) are available. Depending on the scenario either the probe or reference image can be considered to be a trusted image, for instance at a border control the probe is captured live and can be considered as being trusted. Differential detection algorithms have shown promising results for detection of some identity attacks, \eg retouching \cite{Rathgeb-DifferentialDetectionRetouching-ACCESS-2020}, makeup \cite{Rathgeb-MakeupAttackDetection-ACCESS-2020}, and morphed images \cite{Scherhag-FaceMorphingAttacks-TIFS-2020}.

Based on the above, we propose, in this work, a framework for detecting physical and digital identity attacks. In order to exploit identity information and achieve high generalisation, a differential approach for anomaly detection is proposed. Firstly, feature embeddings are obtained from a suspected and a trusted image. After that, the extracted information is fused and given as input to a one-class classifier. Several anomaly detection techniques are evaluated in the experimental evaluation carried out over several databases containing different digital and physical identity attacks. To sum up, this work makes the following contributions:  

\begin{itemize}
    \item A differential anomaly detection framework for unknown face identity attacks.
    \item An extensive evaluation of the generalisability of the proposed framework across multiple identity attack types in both the digital and physical domain.
    \item A highly generalisable framework for detecting digital and physical identity attacks, trained on only bona fide images.
\end{itemize}

The remainder of the paper is organised as follows: Sect.~\ref{sec:related_work} briefly mentions related works. An overview of the proposed framework is shown in Sect.~\ref{sec:proposed_framework}. Sect.~\ref{sec:experimental_evaluation} describes the experimental setup including used databases and metrics. Sect.~\ref{sec:results} reports the experimental results of the proposed framework on several attack types in both the physical and digital domains. Finally, Sect.~\ref{sec:conclusion} concludes the paper with a summary of the obtained results and accomplishments.

\section{Related Work}
\label{sec:related_work}
Many previous works focus on detecting a single or a few related attack types. A lot of attention has been on detecting APs, especially print and replay attacks \cite{Raghavendra-FacePAD-Survey-2017}. Later, researchers showed the vulnerability of face recognition systems to digital manipulations \cite{Ferrara-TheMagicPassport-IJCB-2014, TolosanaDeepFakesBeyondSurveyOfFaceManipulationFaceDetection2020} which have gained much attention in recent years. In particular, DeepFakes and morphing attacks pose severe challenges, as the former can be used to spread misinformation, and morphed images can be used to bypass automated face recognition systems. 

Regarding Presentation Attack Detection (PAD), several hardware-based approaches have been proposed to detect APs in the physical domain, \eg based on reflection, thermal radiation, and motion estimation. Those approaches are usually tailored towards detecting specific attack types or Presentation Attack Instrument (PAI) species and often require specific and expensive sensors. In contrast to hardware-based methods, software-based techniques have been proposed to spot attacks in the physical and digital domain. In general, the existing detection schemes use $i)$ texture analysis~\cite{GonzalezSoler-UnkownAttacksFace-IET-Biometrics-2021}, $ii)$ digital forensics~\cite{Rathgeb-PRNU-Retouching-Detection-BMT-2020}, or $iii)$ deep-learning techniques~\cite{Wang-PAD-AdvDomainAdaptation-TIFS-2020,Deb-UniFAD-Arxiv-2021}. 

In addition, there exist some studies which have focused on the detection of multiple attack types, for instance, FaceGuard \cite{Deb-FaceGuard-arxiv-2020} which obtained a $99.81\%$ detection accuracy on unknown adversarial attacks generated by six different tools. Most existing works in this category focus on the detection of attacks within the same domain, \eg detecting physical attacks, and only a few works have proposed solutions to generalise to attacks in both the physical and digital domain. Mehta \etal~\cite{Mehta-CraftingAPanopticFacePresentationAttackDetector-ICB-2019} proposed an algorithm that showed promising detection results on three PAI species in the physical domain (silicone mask, photo-, and video-replay attacks) and one attack in the digital domain (face swap). In \cite{Deb-UniFAD-Arxiv-2021}, Deb \etal proposed a multi-task learning framework with k-means clustering, which showed high detection accuracy $({\sim}94.73\%)$ on a database comprising 25 attack types across three different attack categories (adversarial, digital, and physical).

\begin{figure}[t!]
    \centering
    \includegraphics[width=\linewidth]{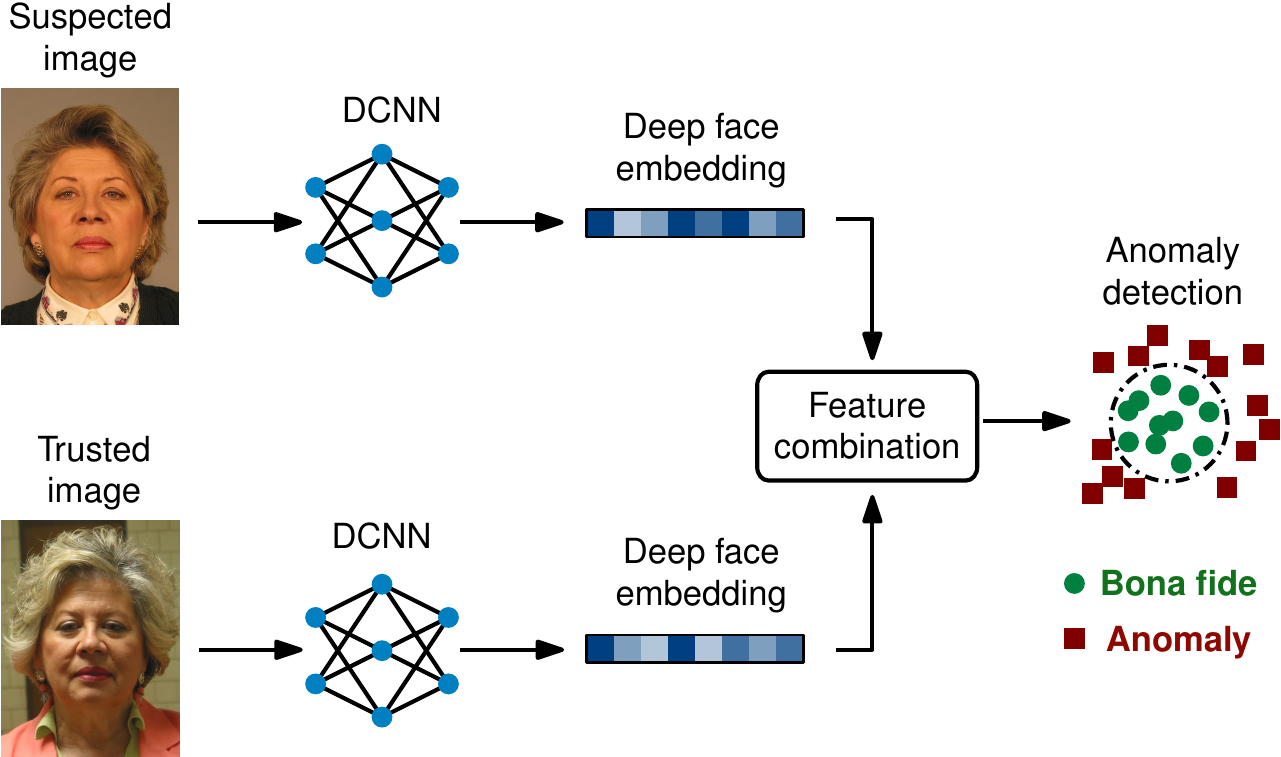}
    \caption{Overview of proposed differential anomaly detection framework.}
    \label{fig:proposed_framewrk}
\end{figure}

\section{Proposed Framework}
\label{sec:proposed_framework}
The proposed framework is inspired by the concept of differential attack detection that was firstly proposed by Scherhag \etal~\cite{Scherhag-FaceMorphingAttacks-TIFS-2020} for morphing attack detection. In said method deep face embeddings were extracted from image pairs and combined as input to a binary classifier. The concept proposed by Scherhag has also been successfully applied to detect retouching \cite{Rathgeb-DifferentialDetectionRetouching-ACCESS-2020} and makeup \cite{Rathgeb-MakeupAttackDetection-ICPR-2020} attacks. However, instead of a binary classifier, the proposed framework implements an anomaly detection module which represents a fundamental difference. An overview of the proposed framework is given in Fig.~\ref{fig:proposed_framewrk}. Two images are given as input, whereafter deep face embeddings are extracted from both images, and the information is fused. The resulting features are fed to an anomaly detection module which classifies the input at hand as being a BP or an anomaly. As it was mentioned, the anomaly detection module is trained using only BPs. The idea behind this is to learn the natural changes (\ie, intra class variation), which can occur, between two BPs of the same subject, \eg changes due to ageing, illumination, and pose. Unnatural and extreme changes not observed on the BPs would be considered as identity attacks. It is expected that the proposed framework will work well on makeup impersonation and morphing attacks as differences in identity between the compared images will be contained in the extracted deep face embeddings. However, it should not appropriately work on all PAI species such as replay and print attacks where the attack does not change the facial appearance and information about the attack is unlikely to be reflected in the extracted deep face embeddings. Similarly, for digital manipulations it is only expected to work in cases where a significant amount of information about the manipulation is stored in the combined feature embedding of the trusted and suspected image. 

\begin{figure}[t!]
    \centering %
\rotatebox[origin=c]{90}{\textsf{Subject 2}}\quad
\begin{subfigure}{0.25\linewidth}
  \includegraphics[width=\columnwidth]{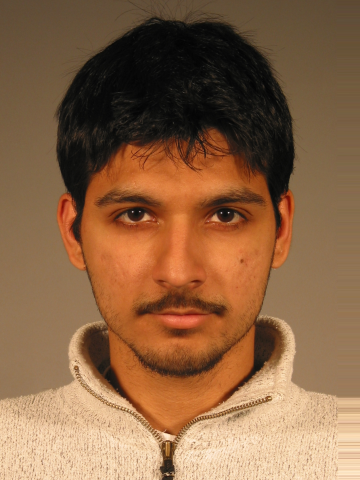}

\end{subfigure}\quad %
\begin{subfigure}{0.25\columnwidth}
  \includegraphics[width=\columnwidth]{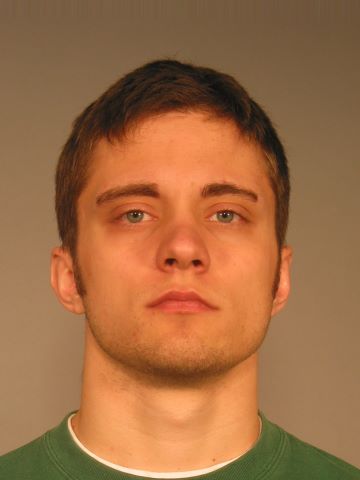}
\end{subfigure}\quad %
\begin{subfigure}{0.25\linewidth}
    \includegraphics[width=\columnwidth]{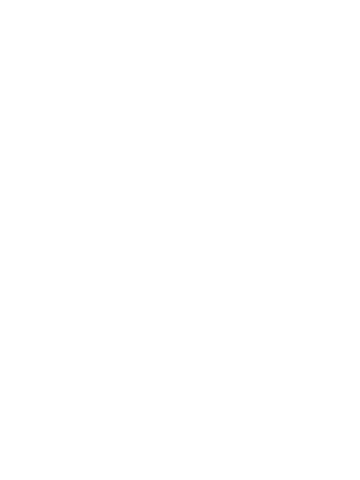}
\end{subfigure} \\
\rotatebox[origin=c]{90}{\textsf{Manipulated}}\quad
\begin{subfigure}{0.25\linewidth}
    \includegraphics[width=\columnwidth]{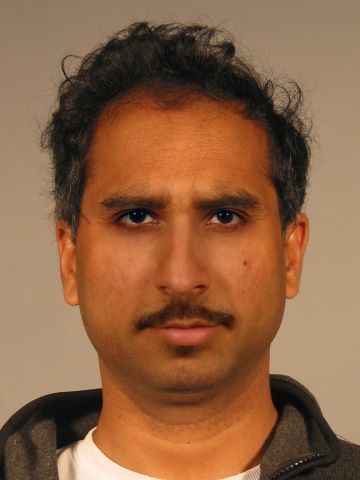}
\end{subfigure}\quad %
\begin{subfigure}{0.25\columnwidth}
  \includegraphics[width=\columnwidth]{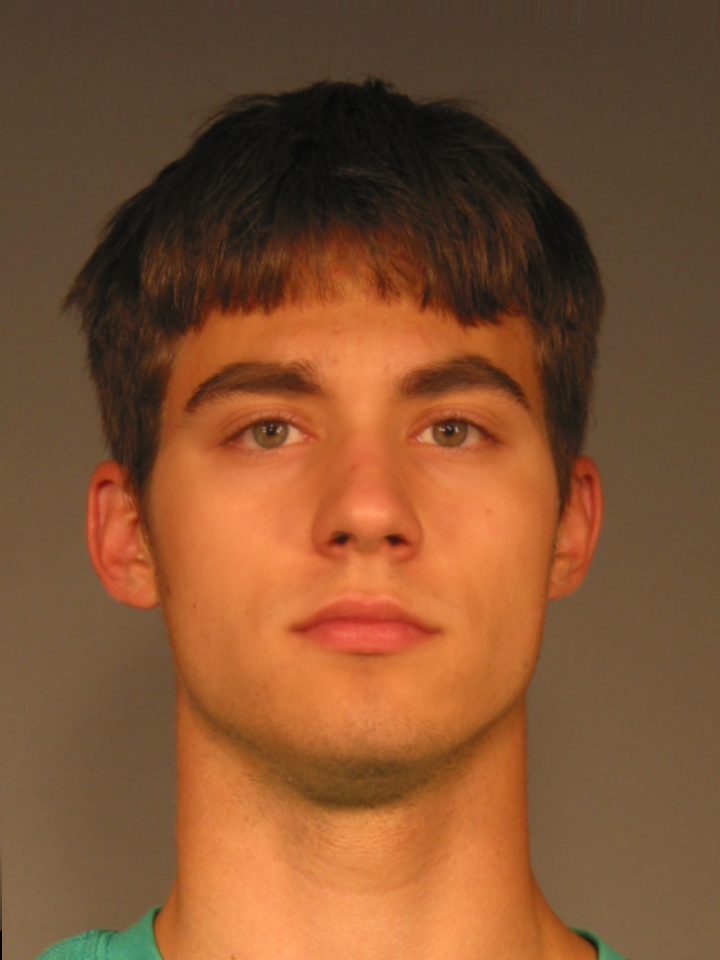}
\end{subfigure}\quad %
\begin{subfigure}{0.25\linewidth}
    \includegraphics[width=\columnwidth]{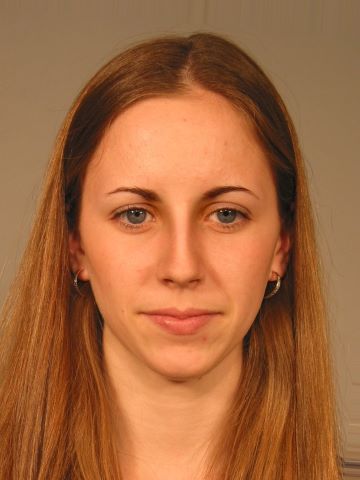}
\end{subfigure} \\
\rotatebox[origin=c]{90}{\textsf{Subject 1}}\quad
\begin{subfigure}{0.25\columnwidth}
   \includegraphics[width=\columnwidth]{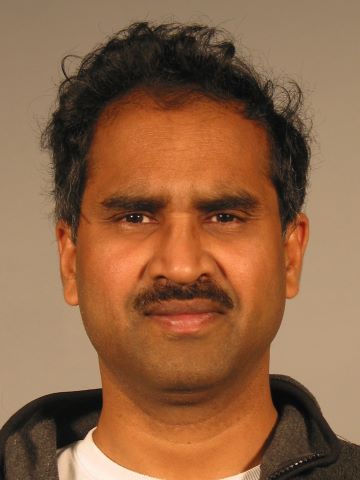}
  \caption{}
\end{subfigure}\quad %
\begin{subfigure}{0.25\columnwidth}
  \includegraphics[width=\columnwidth]{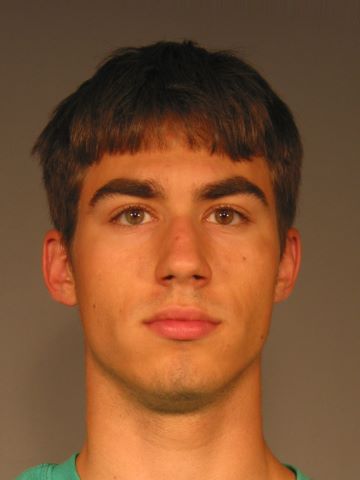}
  \caption{}
\end{subfigure}\quad %
\begin{subfigure}{0.25\columnwidth}
\includegraphics[width=\columnwidth]{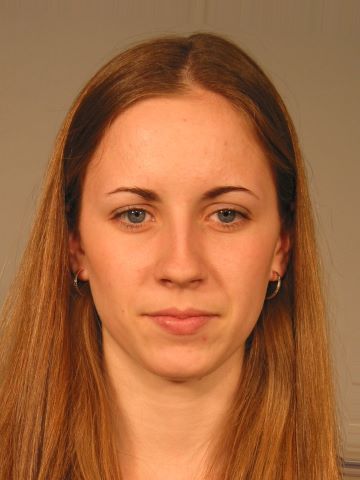}
  \caption{}
\end{subfigure}
 \caption{Examples of manipulated images generated from the FRGC database. (a) face swap, (b) morphing, and (c) retouching.}
 \label{fig:digital_attack_examples}
\end{figure}

For the extraction of deep face embeddings, we propose the use of existing and pre-trained state-of-the-art face recognition models. The advantage of using this approach is that such networks have shown to achieve latent representations with low intra-class and high inter-class variation. For the evaluation in this paper, a pre-trained model of ArcFace~\cite{Deng-ArcFace-IEEE-CVPR-2019} based on ResNet100 is used for extracting deep face embeddings and faces are aligned using RetinaFace~\cite{Deng-RetinaFace-CVPR-2020}. During the evaluation four one-class classifiers (Gaussian Mixture Model (GMM), Support Vector Machines (SVM), Variational Autoencoder (VAE)~\cite{Kingma-VAE-arxiv-2014} \cite{Zhao-Pyod-JournalOfMachineLearningResearch-2019}, Single-Objective Generative Adversarial Active Learning~(SO-GAAL)~\cite{Liu-SOGAAL-arxiv-2019} \cite{Zhao-Pyod-JournalOfMachineLearningResearch-2019}) and three fusion schemes ($\mathrm{SUB}$, $\mathrm{SUB}^2$, $\mathrm{ABS}$) are evaluated. Given two deep face embeddings A (from the reference image) and B (from the probe image):

\begin{equation}
    \mathrm{SUB} = A - B
\end{equation}
\begin{equation}
    \mathrm{SUB}^2 = (A - B)^2
\end{equation}
\begin{equation}
    \mathrm{ABS} = {\mid}A - B{\mid}
\end{equation}

\section{Experimental Evaluation}
\label{sec:experimental_evaluation}

The experimental evaluation addresses the following goals: $i)$ analyse the detection performance of our scheme for different fusion operations and state-of-the-art classifiers, and $ii)$ evaluate exhaustively the best performing pipeline for unknown digital and physical attacks. The experiments do not explicitly consider the scenario where an attacker has prior knowledge of the specific security mechanisms embedded in a face recognition system. It is possible that an attacker can use such information to circumvent the detection provided by our algorithm, thereby gaining unauthorized access. This scenario could be further explored in the future.

\begin{figure}[t!]
    \centering %
\rotatebox[origin=c]{90}{\textsf{TARGET}}\quad
\begin{subfigure}{0.28\linewidth}
  \includegraphics[width=\columnwidth]{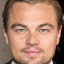}

\end{subfigure}\quad %
\begin{subfigure}{0.28\columnwidth}
\includegraphics[width=\columnwidth]{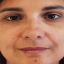}
\end{subfigure}\quad %
\begin{subfigure}{0.28\linewidth}
    \includegraphics[width=\columnwidth]{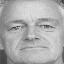}
\end{subfigure} \\
\rotatebox[origin=c]{90}{\textsf{AP}}\quad
\begin{subfigure}{0.28\columnwidth}
   \includegraphics[width=\columnwidth]{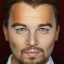}
  \caption{}
\end{subfigure}\quad %
\begin{subfigure}{0.28\columnwidth}
\includegraphics[width=\columnwidth]{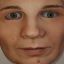}
  \caption{}
\end{subfigure}\quad %
\begin{subfigure}{0.28\columnwidth}
\includegraphics[width=\columnwidth]{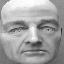}
  \caption{}
\end{subfigure}
 \caption{Examples of physical attacks in each of the used databases. (a) HDA\_MPA\_DB, (b) CSMAD-Mobile, (c) XCSMAD.}
 \label{fig:physical_attack_examples}
\end{figure}

\begin{table*}[t!]
    \centering
\caption{Detection equal error rate (D-EER \%) for the different models and fusion schemes on the used evaluation databases.}
\begin{adjustbox}{max width=\textwidth}
   \begin{tabular}{@{\extracolsep{2pt}}llrrrrrrrrrrrrrr@{}} \toprule 
       & & \multicolumn{4}{c}{\textbf{FERET}}  &
      \multicolumn{4}{c}{\textbf{FRGC}} & \multicolumn{3}{c}{\textbf{CSMAD}}  &
      \multicolumn{1}{c}{\textbf{XCSMAD}} & \multicolumn{1}{c}{\textbf{Makeup}} &
      \multicolumn{1}{c}{\textbf{Avg.}} \\  \cmidrule{3-6} \cmidrule{7-10} \cmidrule{11-13} \cmidrule{14-14} \cmidrule{15-15} \cmidrule{16-16}
\textbf{Model} & \textbf{Fusion}  &  \multicolumn{1}{c}{Swap Outer} & \multicolumn{1}{c}{Swap Inner} & \multicolumn{1}{c}{Morphing} & \multicolumn{1}{c}{Retouching} & \multicolumn{1}{c}{Swap Outer} & \multicolumn{1}{c}{Swap Inner} & \multicolumn{1}{c}{Morphing} & \multicolumn{1}{c}{Retouching} & \multicolumn{1}{c}{iPhone} & \multicolumn{1}{c}{S7} & \multicolumn{1}{c}{S8}  \\
\midrule
   \multirow{3}{*}{GMM} & SUB & $0.00$ & $0.19$ & $2.26$ & $17.71$ & $0.56$ & $0.70$ & $5.69$ & $24.64$ & $0.00$ & $0.00$ & $0.00$ & $0.00$ & $3.56$ & $4.25$  \\
                        & SUB$^2$  & $0.00$ & $0.25$ & $2.26$ & $17.74$ & $0.50$ & $0.76$ & $5.94$ & $24.64$ & $0.00$ & $0.00$ & $0.00$ & $0.00$ & $3.56$ & $4.28$ \\
                        & ABS & $0.00$ & $0.25$ & $2.26$ & $18.98$ & $0.55$ & $0.70$ & $5.57$ & $24.18$ & $0.00$ & $0.00$ & $0.00$ & $0.00$ & $3.56$ & $4.31$ \\ \midrule
   \multirow{3}{*}{SVM} & SUB & $0.00$ & $0.15$ & $2.26$ & $17.74$ & $0.56$ & $0.70$ & $5.69$ & $24.70$ & $0.00$ & $0.00$ & $0.00$ & $0.00$ & $3.56$ & $4.26$\\
                        & SUB$^2$  & $0.00$ & $0.25$ & $2.26$ & $17.83$ & $0.50$ & $0.76$ & $5.96$ & $24.70$ &  $0.00$ & $0.00$ & $0.00$ & $0.00$ & $3.56$ & $4.29$\\
                        & ABS & $0.00$ & $0.25$ & $2.41$ & $23.34$ & $0.53$ & $0.71$ & $5.64$ & $24.15$ & $0.00$ & $0.00$ & $0.00$ & $0.00$ & $3.56$ & $4.66$\\ \midrule
   \multirow{3}{*}{VAE} & SUB & $0.00$ & $0.19$ & $2.20$ & $17.71$ & $0.56$ & $0.70$ & $5.60$ & $24.49$ &$0.00$ & $0.00$ & $0.00$ & $0.00$ & $3.56$ & \textbf{4.23}\\
                        & SUB$^2$  & $0.00$ & $0.25$ & $2.26$ & $17.74$ & $0.50$ & $0.74$ & $6.10$ & $24.55$ &$0.00$ & $0.00$ & $0.00$ & $0.00$ & $3.56$ & $4.28$\\
                        & ABS & $0.00$ & $0.25$ & $2.10$ & $18.85$ & $0.56$ & $0.68$ & $5.67$ & $24.01$ &$0.00$ & $0.00$ & $0.00$ & $0.00$ & $3.56$ & $4.28$\\ \midrule
\multirow{3}{*}{SO-GAAL}& SUB & $1.19$ & $1.66$ & $5.66$ & $19.39$ & $5.49$ & $5.29$ & $14.47$ & $25.21$ & $0.00$ & $0.00$ & $0.00$ & $3.02$ & $13.47$ & $7.30$\\
                        & SUB$^2$ & $0.00$ & $0.25$ & $2.26$ & $17.71$ & $0.71$ & $0.89$ & $5.63$ & $24.76$ &$0.00$ & $0.00$ & $0.00$ & $0.00$ & $3.56$ & $4.29$\\
                        & ABS & $19.62$ & $20.48$ & $22.28$ & $39.49$ & $31.36$ & $31.83$ & $35.88$ & $46.39$ & $5.45$ & $10.75$ & $14.47$ & $35.04$ & $26.78$ & $26.14$\\
        \bottomrule
    \end{tabular}
\label{tab:deer_all_approaches}
\end{adjustbox}{}
\end{table*}

\subsection{Databases}

The  experimental  evaluation  was  conducted  over  several databases. For the identity attack detection, one of the input samples is always considered as a BP whilst the other image could be either an identity attack or a BP.

\subsubsection{Bona Fide Presentations}
For training the proposed framework, the academic version of the UNCW MORPH \cite{uncw-morph} database and the CASIA-FaceV5 \cite{casia-facev5} database are used. As the UNCW MORPH database contains few subjects of Asian ethnicity, images from the CASIA-FaceV5 database are also used. Furthermore, since the proposed framework requires paired images, we for each subject select all possible unique pairs of images of that subject. These pairs of images are used during training. Additionally, pairs of bona fide images from the FERET~\cite{Phillips-FERET-1998}, FRGCv2~\cite{Phillips-FRGC-2005}, XCSMAD~\cite{Kotwal-MultispectralDeepEmbeddingsCountermeasureToSiliconeMaskPAs-TBIOM-2019}, CSMAD-Mobile~\cite{Raghavendra-CustomSiliconFaceMask-VulnerabilityAndPAD-IWBF-2019}, and HDA\_MPA\_DB~\cite{Drozdowski-MakeupPADataset-IWBF-2021} are used during testing.

\subsubsection{Digital Manipulations}
To evaluate the efficacy of the proposed algorithm for detecting digital manipulations, a database compromising three common digital manipulations (retouching, morphing, and face swapping) is used. The images have been created from a subset of images from the FERET \cite{Phillips-FERET-1998} and FRGCv2 \cite{Phillips-FRGC-2005} database which prior to the application of the above manipulations have been normalised. For each type of manipulation two tools have been used. For retouching\footnote{Retouched images are also referred to as photoshopping and concerns digital alterations of a face, \eg for beautification}, \textit{InstaBeauty}~\cite{instabeauty} and \textit{Fotorus} \cite{fotorus}, for morphing \textit{FaceFusion}~\cite{facefusion} and \textit{UBO Morpher}~\cite{bologa_biotrec_lab, Ferrara-TextureBlendingAndShapeWarpingInFaceMorphing-IEEE-BIOSIG-2019} are used, whereas for face swapping~\textit{fewshot-face} \cite{fewshot-face} and \textit{simple\_faceswap}~\cite{simple-faceswap} are used. Examples of each type of manipulation is shown in Fig.~\ref{fig:digital_attack_examples}. As can be seen, face swapping swaps the face of subject 2 onto subject 1, morphing combines the facial attributes of the two subjects and retouching slightly alters the subject's face by, for instance, a slimming of the nose and enlargement of the mouth. Note that for retouching only the image of a single subject is required.

For the evaluation of swapped images, two scenarios referred to as \textit{face swap inner} and \textit{face swap outer} are employed:

\begin{itemize}
    \item \textit{face swap outer}, a probe image of the individual contributing to the outer part of the image (the source) is employed during detection.
    \item \textit{face swap inner}, a probe image of the target identity is used.
\end{itemize}

For morphing, we only evaluate the scenario where a probe image of the individual contributing to the outer part of the morphed image (the source) is used.

\subsubsection{Attack Presentations}
For evaluating the efficacy of the proposed framework towards physical identity attacks, three databases are used: the \textit{XCSMAD} \cite{Kotwal-MultispectralDeepEmbeddingsCountermeasureToSiliconeMaskPAs-TBIOM-2019} and \textit{CSMAD-Mobile} \cite{Raghavendra-CustomSiliconFaceMask-VulnerabilityAndPAD-IWBF-2019} database comprising of custom silicone mask attacks as well as a subset of the Hochschule Darmstadt (HDA) facial makeup presentation attack database (HDA\_MPA\_DB) \cite{Drozdowski-MakeupPADataset-IWBF-2021} consisting of bona fide makeup and impersonation attacks. Examples from each database are shown in Fig.~\ref{fig:physical_attack_examples}.

\subsection{Experimental Metrics}
The proposed framework is evaluated empirically in compliance with ISO/IEC 30107-Part 3~\cite{ISO-IEC-30107-3-PAD-metrics-170227} for biometric PAD. Specifically, we report:

\begin{itemize}
    \item Attack Presentation Classification Error Rate (APCER), which is the proportion of attack presentations or identity attacks misclassified as bona fide presentations.
    \item Bona Fide Presentation Classification Error Rate (BPCER), which is the proportion of bona fide presentations wrongly classified as attack presentations.
\end{itemize}

Building upon these metrics, we also report: $i)$ Detection Error Tradeoff
(DET) curves between APCER and BPCER; $ii)$ the BPCERs observed at different APCER values or security thresholds such as at 1\% (BPCER100) and $iii)$ the Detection Equal Error Rate (D-EER), \ie the point where APCER and BPCER are equal.

\begin{figure}[t!]
\begin{subfigure}[t]{0.48\linewidth}
    \centering
  \includegraphics[width=\linewidth]{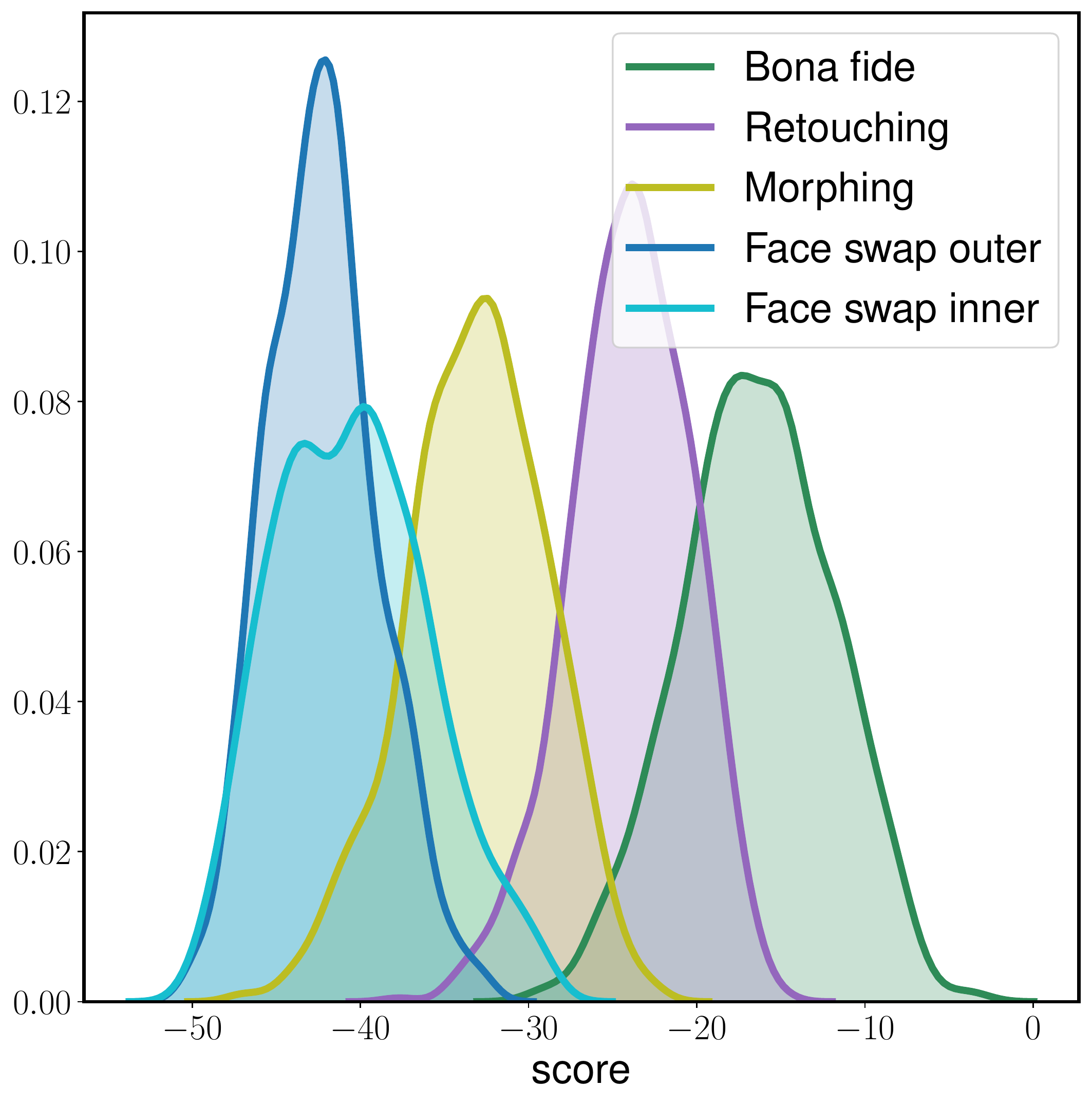}
  \caption{FERET}
\end{subfigure}\quad %
\begin{subfigure}[t]{0.48\linewidth}
    \centering
  \includegraphics[width=\linewidth]{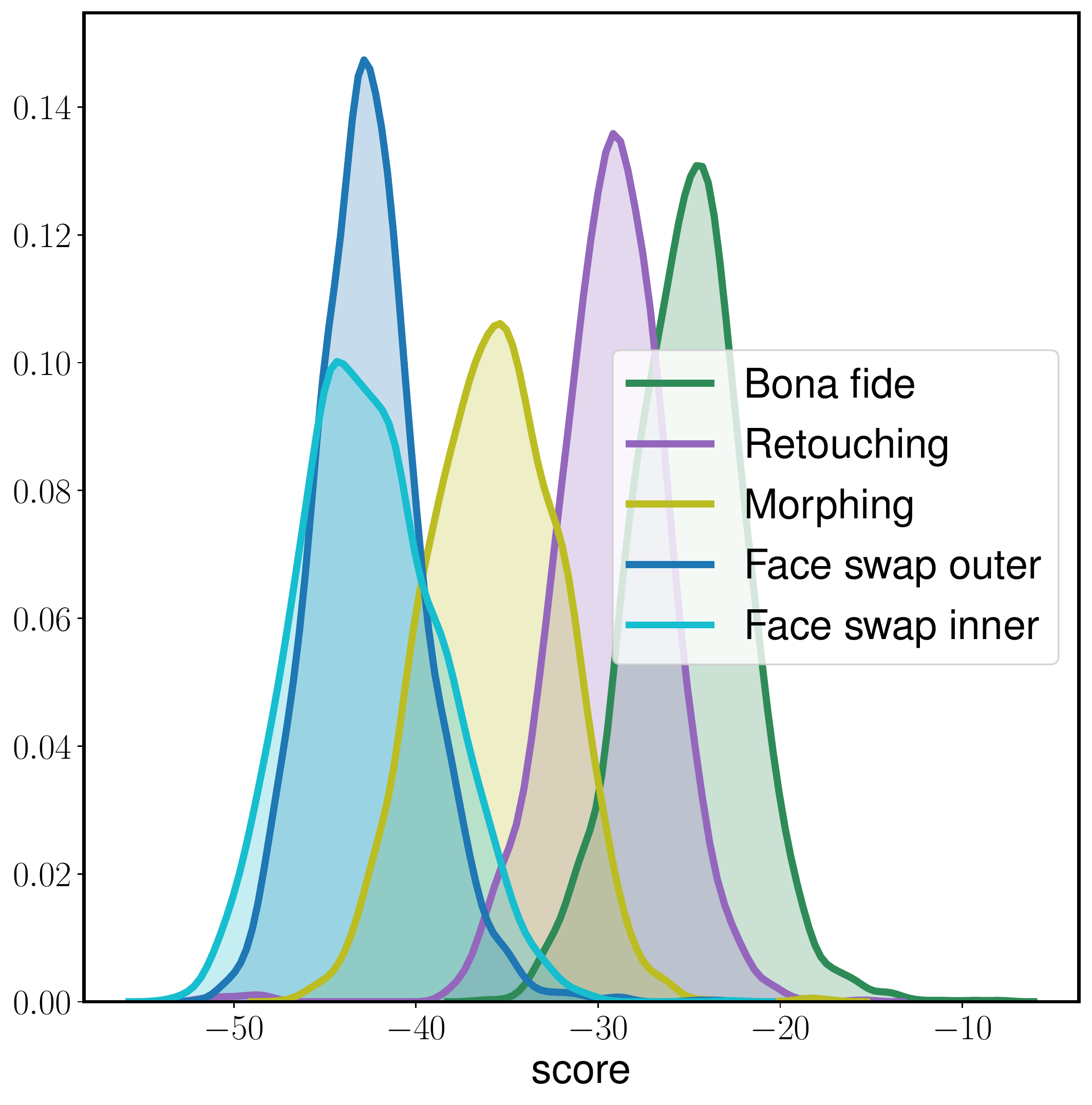}
  \caption{FRGC}
\end{subfigure}\quad %
\caption{Score distributions for the VAE model with the $\mathrm{SUB}$ fusion scheme on bona fide and digitally manipulated images.}
\label{fig:digital_kde_plot}
\end{figure}

\section{Results and Discussion}
Tab.~\ref{tab:deer_all_approaches} shows the D-EERs for different configurations of the proposed framework on three types of digital manipulations and two different PAI species. The results in the table show that the performance of three out of four models over the fusion schemes are very similar, thereby resulting in a D-EER ranging 0.0\% - 24.70\% depending on the type of identity attack. In particular, VAE attains its best detection performance for the $\mathrm{SUB}$ fusion scheme: D-EERs close to 0.0\% for most attacks types and a mean D-EER of 4.23\% show the soundness of our fused representation. Based on the above, we therefore consider the VAE and $\mathrm{SUB}$ fusion scheme for the rest of the evaluation. A comparison of the proposed method to the state-of-the-art is deliberately avoided as most existing methods require knowledge of the attacks during training or are not identity-aware, which would make a comparison unfair and misleading.

\begin{figure}[t!]
\begin{subfigure}[t]{0.48\linewidth}
    \centering
  \includegraphics[width=\linewidth]{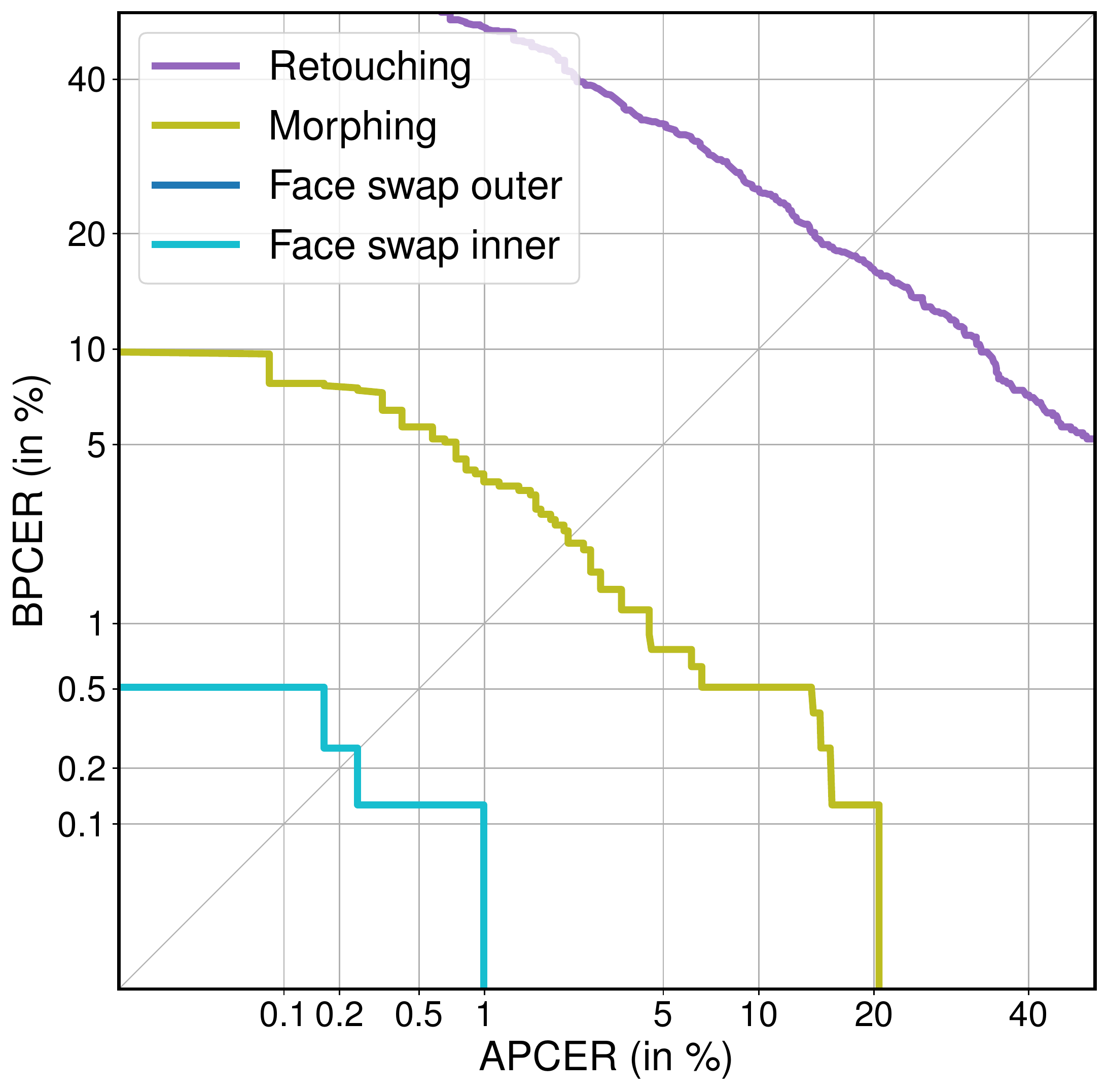}
  \caption{FERET}
\end{subfigure}\quad %
\begin{subfigure}[t]{0.48\linewidth}
    \centering
  \includegraphics[width=\linewidth]{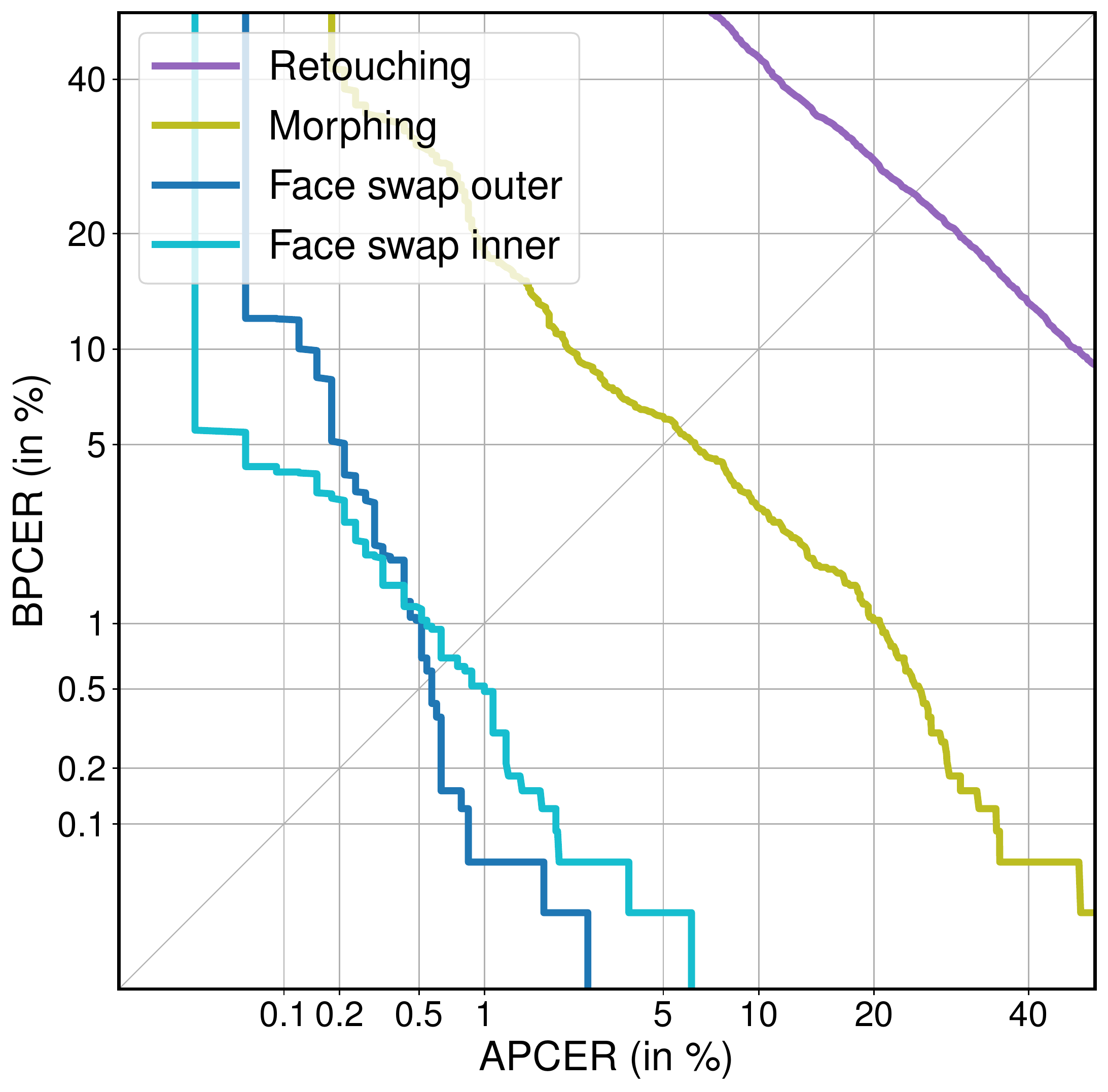}
  \caption{FRGC}
\end{subfigure}\quad %
\caption{DET-curves for the VAE model with the $\mathrm{SUB}$ fusion scheme on bona fide and digitally manipulated images. For (a) face swap outer attain a BPCER = 0.0\% for any APCER; hence, its corresponding curve is not shown.}
\label{fig:digital_det_curve}
\end{figure}

\label{sec:results}
\subsection{Analysis on Digital Manipulations}
The score distributions obtained for the selected configuration of the proposed framework and their corresponding DET-curves on the digital manipulations are shown in Fig.~\ref{fig:digital_kde_plot} and \ref{fig:digital_det_curve}, respectively. As it can be noted, our proposed framework can successfully generalise to several attacks and achieves relatively low detection errors on the swapped and morphed images, especially for the FERET database. In particular, a BPCER in the range of 0.0\% to 17.92\% is reported for an APCER of 1.0\% over both databases. 

In addition, we can observe a poor detection performance over retouched images (\ie, a BPCER greater than 40.0\% for an APCER of 1.0\%). These results are to be expected, as the tools employed in the creation of retouched images alter the appearance of the facial images only moderately. Moreover, the algorithms aim at beautification and not at changing the facial identity. Finally, for the defined face swapping scenarios, the results show a high detection performance, with a BPCER of $0.0\%$ to $0.49\%$ for an APCER of 1.0\%.

\subsection{Analysis on Attack Presentations}
\label{sec:evaluation_physical}
\begin{figure}[t!]
\begin{subfigure}[t]{0.48\linewidth}
    \centering
  \includegraphics[width=\linewidth]{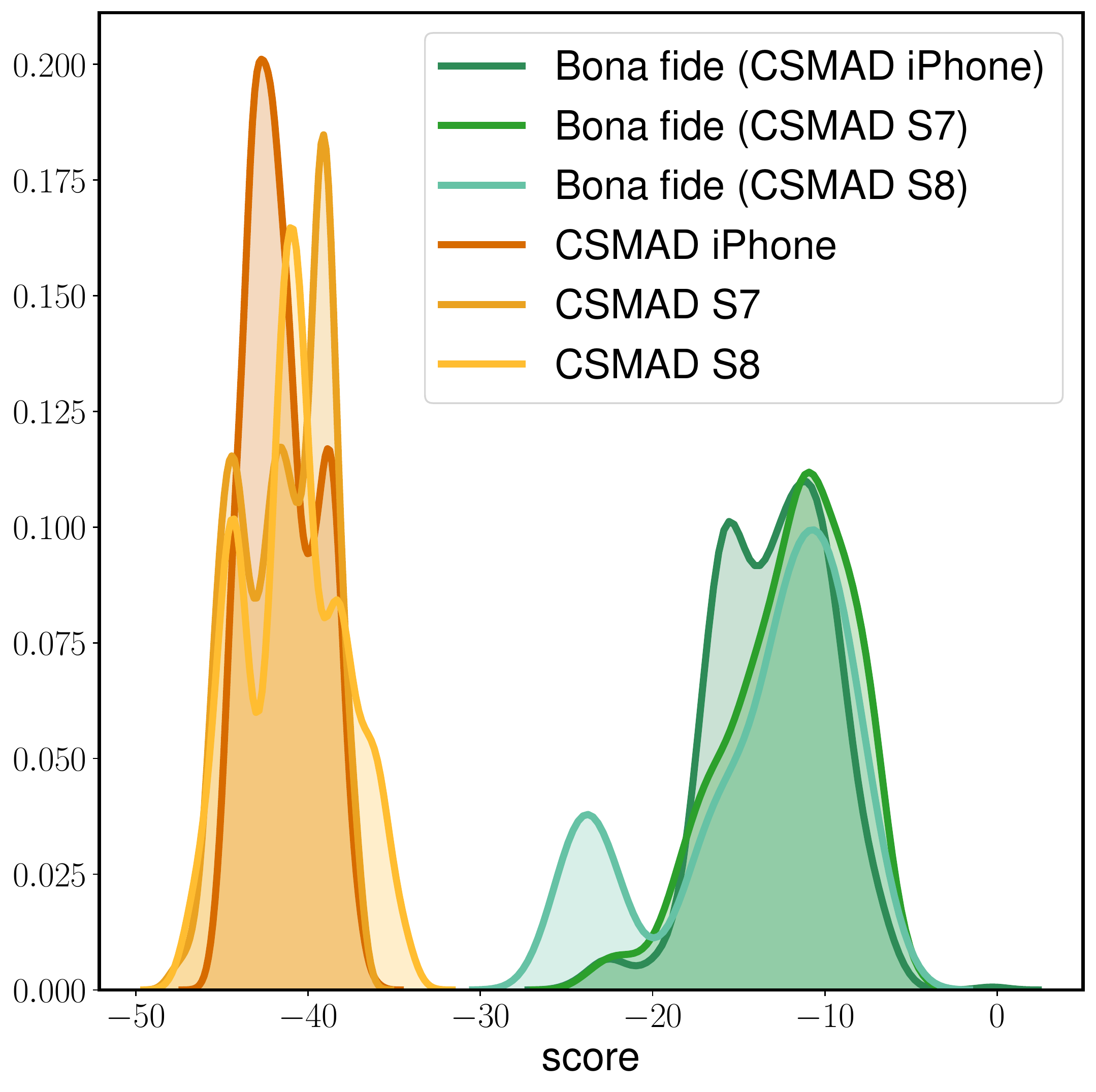}
\end{subfigure}\quad %
\begin{subfigure}[t]{0.48\linewidth}
    \centering
  \includegraphics[width=\linewidth]{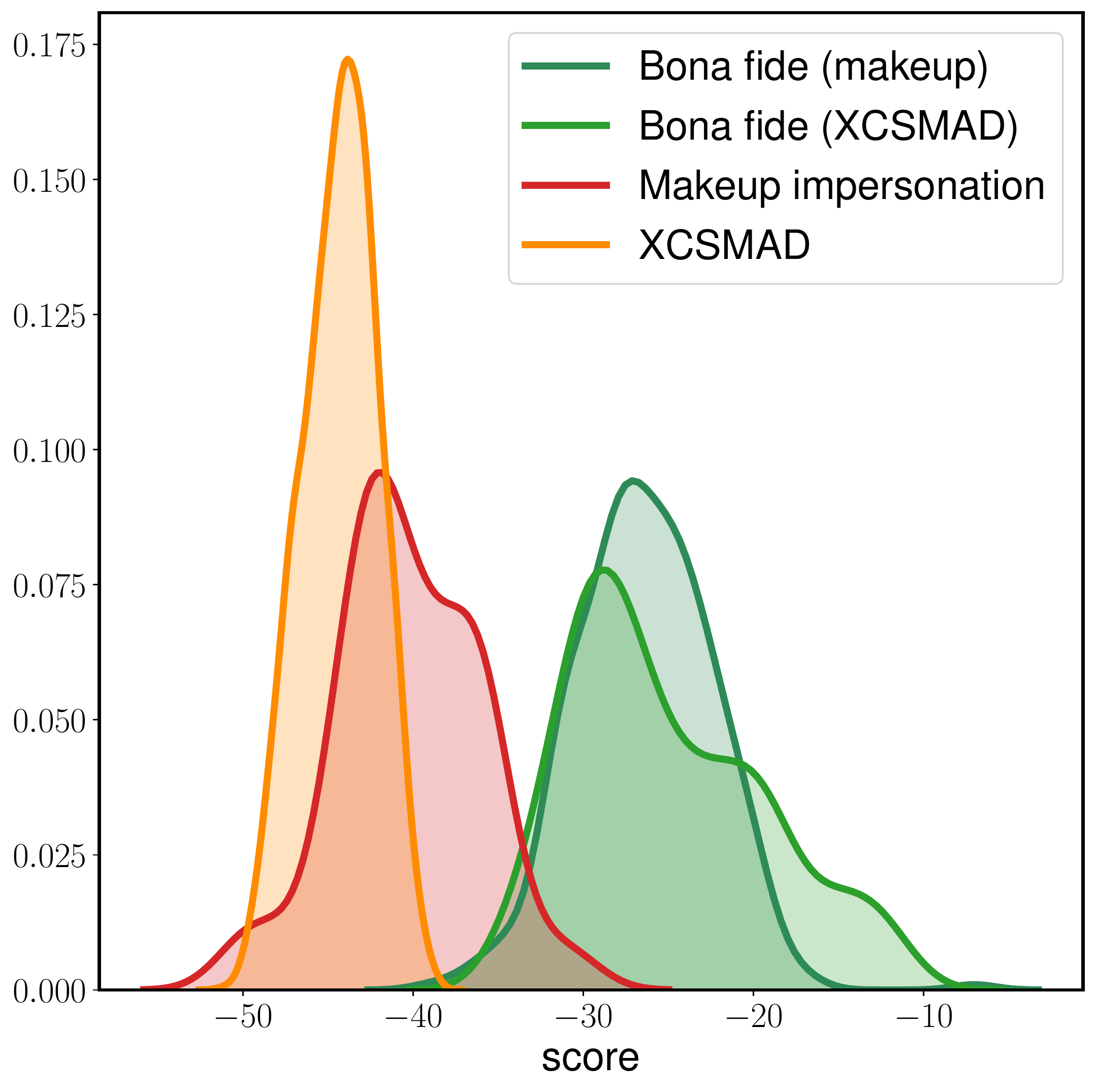}
\end{subfigure}\quad %
\caption{Scores obtained on the different physical databases using VAE with the $\mathrm{SUB}$ fusion scheme.}
\label{fig:physical_kde_plot}
\end{figure}

\begin{figure}[t!]
    \centering
  \includegraphics[width=0.65\linewidth]{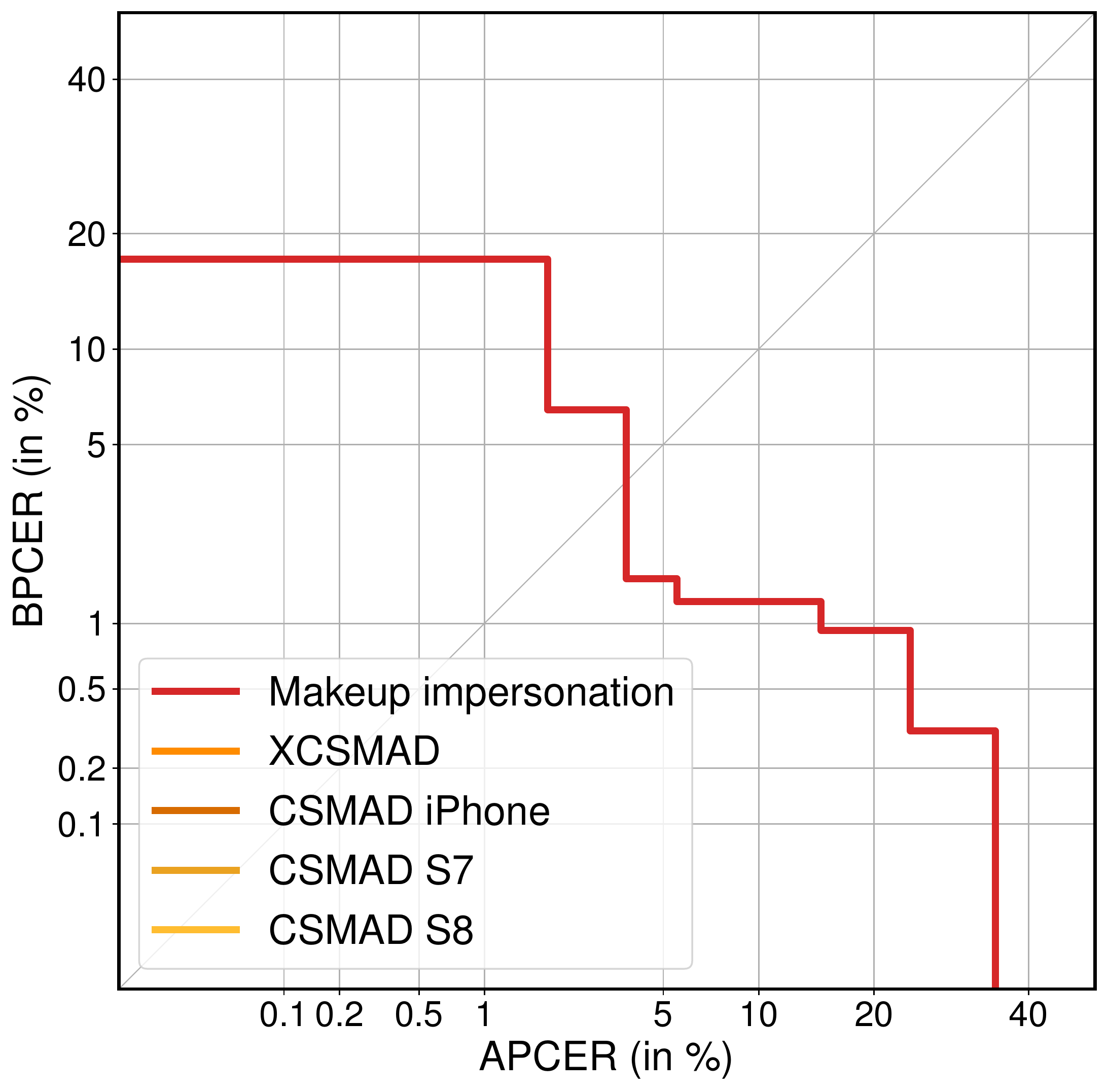}
  \caption{DET-curve for physical attacks. Plots which are not shown obtain a D-EER of 0\%.}
\label{fig:physical_det_curve}
\end{figure}

We also evaluate our proposed scheme for attacks in the physical domain. Fig.~\ref{fig:physical_kde_plot} shows the score distributions obtained for the different PAIs and corresponding BPs. The results indicate that the BPs can be successfully separated from their corresponding APs, and only an overlap between the makeup impersonation attacks and makeup bona fide scores can be perceived. Consequently, similar results are reported in Fig.~\ref{fig:physical_det_curve}: a BPCER of 0.0\% for most PAI species at a APCER of 1.0\% does confirm the soundness of our proposed framework to detect unknown PAIs. It should be noted that for the physical attacks a relative low number of images have been used and the results should be interpreted with care.

\begin{figure*}[t!]
\begin{subfigure}[t]{0.24\linewidth}
  \includegraphics[width=\linewidth]{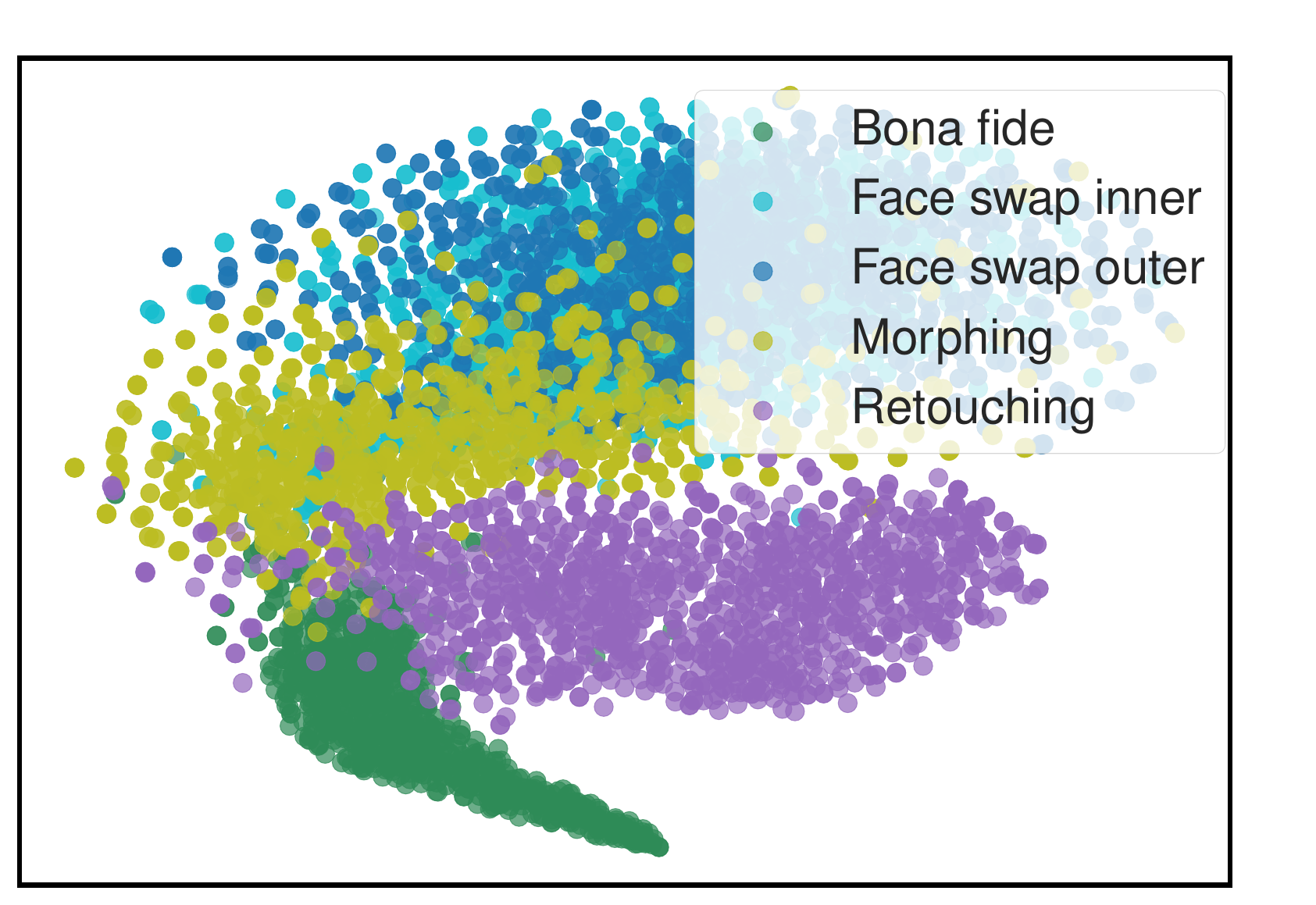}
  \caption{FERET Digital}
\end{subfigure} %
\begin{subfigure}[t]{0.24\linewidth}
  \includegraphics[width=\linewidth]{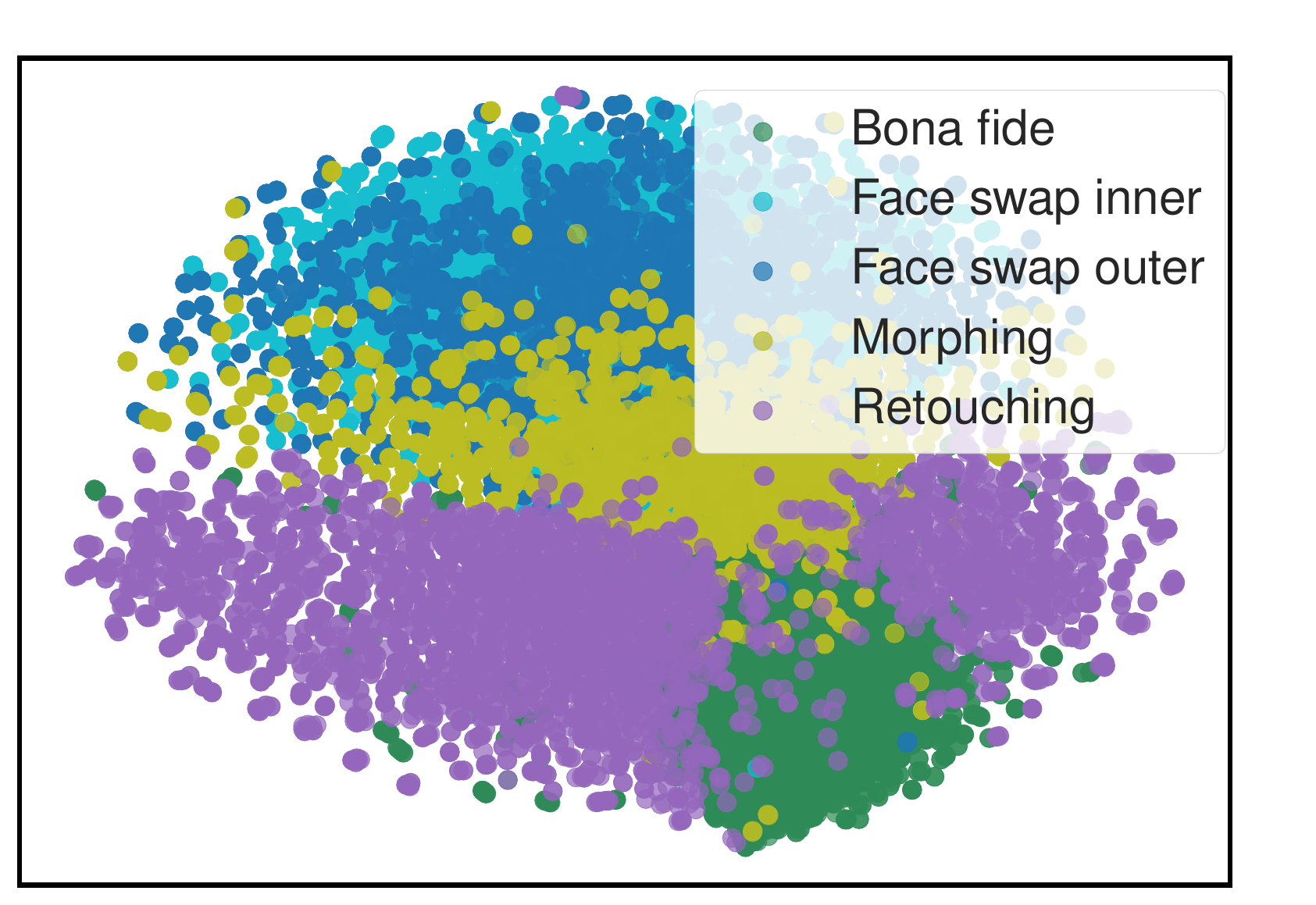}
  \caption{FRGC Digital}
\end{subfigure} %
\centering
\begin{subfigure}[t]{0.24\linewidth}
  \includegraphics[width=\linewidth]{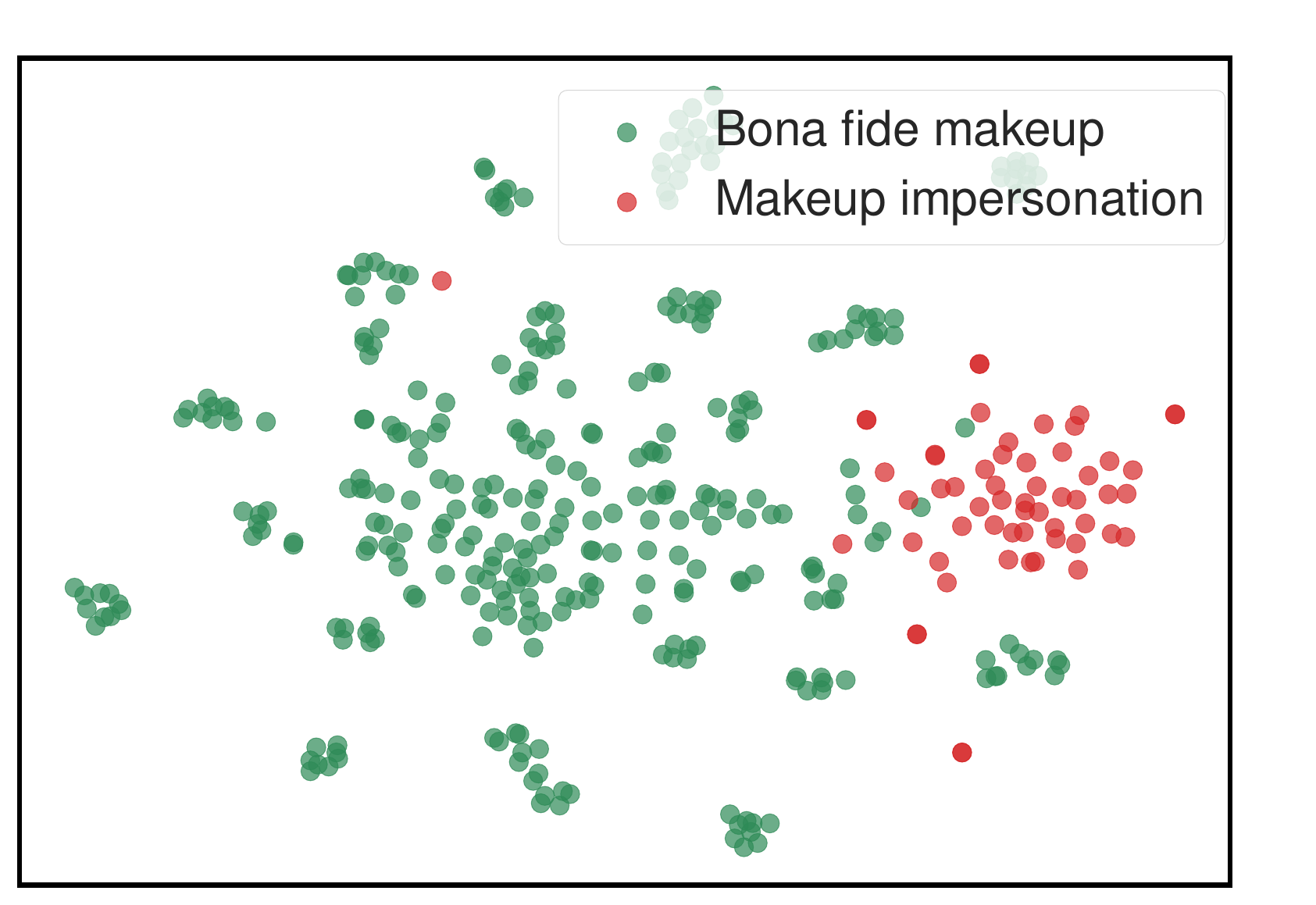}
  \caption{Makeup Impersonation}
\end{subfigure} %
\begin{subfigure}[t]{0.24\linewidth}
  \includegraphics[width=\linewidth]{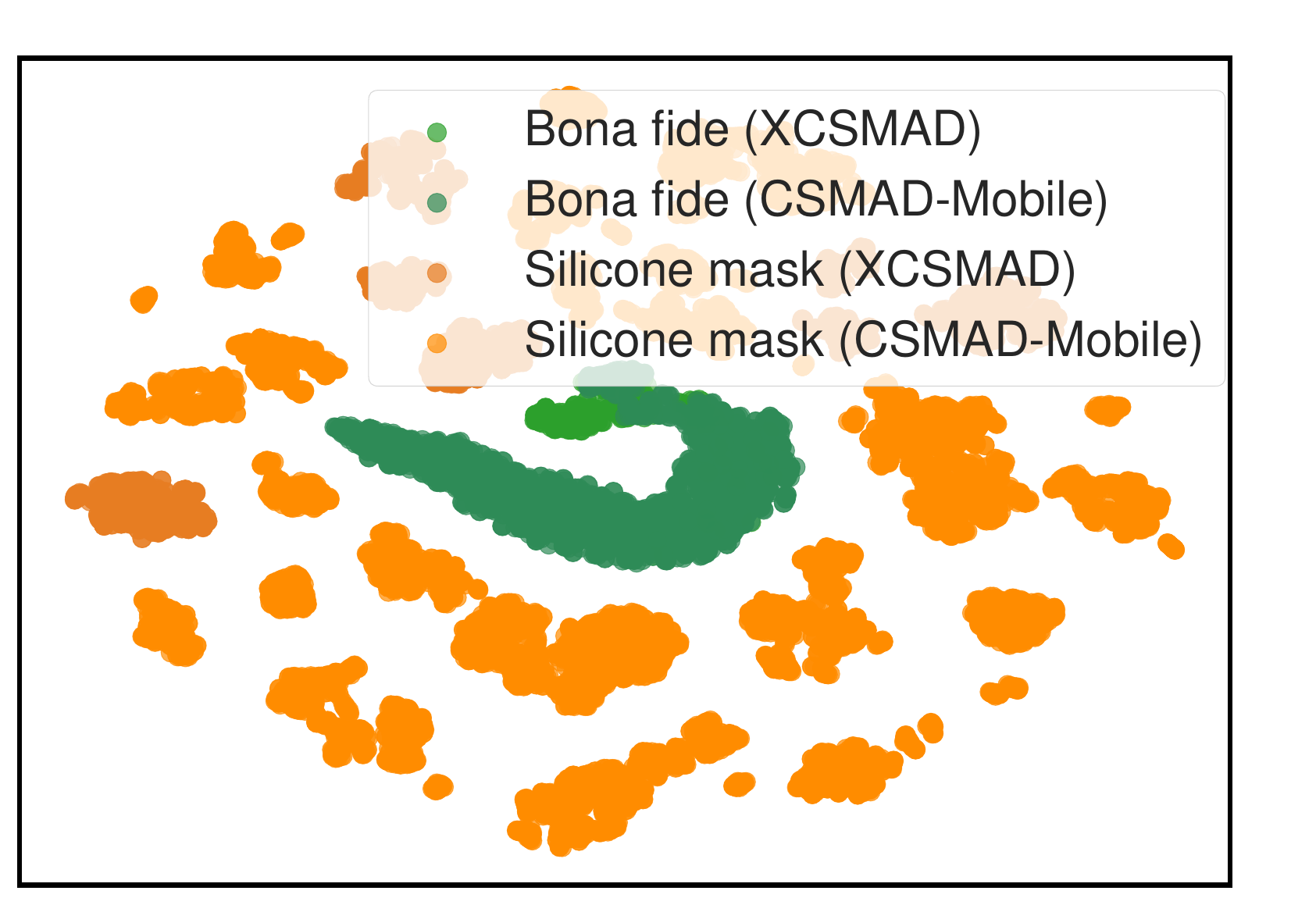}
  \caption{Masks}
\end{subfigure} %
\caption{t-SNE visualisation of the ArcFace face embeddings after applying the $\mathrm{ABS}$ fusion scheme.}
\label{fig:tsne-figures}
\end{figure*}

Finally, Fig.~\ref{fig:tsne-figures} visualises the t-SNE plots for the deep face embeddings used during evaluation for the $\mathrm{ABS}$ fusion scheme. The $\mathrm{ABS}$ fusion scheme was chosen for better visualisation and due to the similar performance (see Tab.~\ref{tab:deer_all_approaches}). The t-SNE plots indicates that it might be possible to separate the embeddings extracted for identity attacks from corresponding embeddings extracted from BP images. The results obtained in the evaluation confirms this observation for the used databases as it is possible to get low detection errors on most of the used types of identity attacks.

\section{Conclusion}
\label{sec:conclusion}
In this paper, a unified framework for the detection of identity attacks in the physical and digital domains was proposed. The suggested framework uses a differential anomaly detection approach where a trusted image is employed together with a suspected image. For attack detection, feature embeddings are extracted from both images, whereafter a fusion scheme is applied, and the resulting feature vector is given as input to a one-class classifier. Said classifier, is trained using only BP images. The proposed method determines whether a suspected image is bona fide or an anomaly. The results show a high generalisation capability and good detection performance on attacks where an individual's identity is significantly changed. In particular, our proposed approach reported a low BPCER100 close to 0.0\% for most PAI species and manipulated images, with the exception of especially the retouched images. Our proposed framework attained a BCPER100 higher than 40.0\% on retouched images, since they were only designed to subtly alter the facial attributes of the face, not to circumvent biometric systems.

\section{Acknowledgement}
This research work has been partially funded by the German Federal Ministry of Education and Research and the Hessian Ministry of Higher Education, Research, Science and the Arts within their joint support of the National Research Center for Applied Cybersecurity ATHENE, the European Union's Horizon 2020 research and innovation programme under the Marie Sk\l{}odowska-Curie grant agreement No. 860813 - TReSPAsS-ETN, and the DFG-ANR RESPECT Project (406880674).

\bibliographystyle{IEEEtran}
\bibliography{bibliography}

\end{document}

%% file: commands.tex
\def\eg{\textit{e.g.}\@\xspace} 
\def\ie{\textit{i.e.}\@\xspace}

\def\etal{\textit{et al.}\@\xspace} 